\documentclass[runningheads]{llncs}
\usepackage{parskip}
\usepackage{graphicx}
\usepackage{hyperref}   
\usepackage{amsmath}
\usepackage{subfig}
\usepackage[table,xcdraw]{xcolor}
\newcommand{\etal}{\textit{et al. }}

\makeatletter
\renewcommand\subsubsection{\@startsection{subsubsection}{3}{\z@}%
    {-\baselineskip}%
    {0.1\baselineskip}%
{\normalfont\normalsize\bfseries}}
\makeatother

\begin{document}
\title{Personalising Digital Health Behaviour Change Interventions using Machine Learning and Domain Knowledge}
\titlerunning{Personalising Digital Health BCIs}
%
 \author{Aneta Lisowska\inst{1} \and
Szymon Wilk\inst{1}\and 
Mor Peleg\inst{2}}
\authorrunning{A.Lisowska et al.}
 \institute{
 Institute of Computing Science, Poznań University of Technology, Poznań, Poland
\email{\{aneta.lisowska, szymon.wilk\}@put.poznan.pl}\\
\and
Department of Information Systems, University of Haifa, Haifa, Israel \\
\email{morpeleg@is.haifa.ac.il}}
\maketitle              
\begin{abstract}
We are developing a virtual coaching system that helps patients adhere to behaviour change interventions (BCI). Our proposed system predicts whether a patient will perform the targeted behaviour and uses counterfactual examples with feature control to guide personalisation of BCI. We use simulated patient data with varying levels of receptivity to intervention to arrive at the study design which would enable evaluation of our system. 
\end{abstract}
%
%

\section{Introduction}

As part of the Horizon 2020 \href{https://capable-project.eu/}{CAPABLE} 
(CAncer PAtient Better Life Experience) project, we are developing a mobile health (mHealth) application that aims to facilitate the mental wellbeing of cancer patients.  It offers multiple evidence-based digital health behaviour change interventions (BCIs) from the domains of mindfulness, physical exercise, and positive thinking, targeting various aspects of wellbeing (e.g., stress, insufficient sleep etc.) \cite{lisowska2023sato}. Each BCI defines a target activity, that when performed regularly may lead to improvement in one or multiple wellbeing outcomes.  For example, evidence-based studies showed that meditation may reduce stress and may improve the quality of sleep.

Potential prolonged wellbeing benefits depend on regular engagement with the recommended activity. To support patients with BCI adherence, we design a Virtual Coaching (VC) system combining knowledge- and data-driven approaches. VC uses personalisation and prediction models. We utilise Behaviour Change Intervention Ontology \cite{michie2020representation} (BCIO) and Fogg’s Behaviour Model \cite{fogg2009behavior} to select modifiable inputs of personalisation model and machine learning to predict patient behaviour. 

Fogg suggests that a person will perform a target behaviour when they are sufficiently motivated, have the skill/ability to perform the behaviour, and there is a trigger prompting them to do it \cite{fogg2009behavior}. 
In our previous work \cite{lisowska2021personalized} we identified some factors influencing these three components, and formulated Fogg's Model as:
\begin{equation}
     Behaviour = \begin{cases}
    1  & \quad \text{if  } (Motivation \times Ability \times Trigger > \text{action threshold } \\
    0  & \quad \text{otherwise}
  \end{cases}
  \label{eq:behaviour}
\end{equation}

Each of Motivation (M), Ability (A) and Trigger (T) components, including the action threshold, varies among individuals and depends on the suggested activity and the user's current context, both internal (e.g., emotional state) and external (e.g., day of the week). Kunzler \etal \cite{kunzler2019exploring} explored factors that influence patient receptiveness to intervention. They discovered that fixed user characteristics (e.g., age, gender, personality) affect general participants' receptiveness, while extrinsic factors (e.g., date, battery status) are associated with receptiveness in the moment. Building upon this distinction, we map general intervention receptiveness to the action threshold and receptiveness in the moment to the trigger magnitude in Fogg's model (refer to the right part of Figure \ref{personalisationSystem2}). It is important to note that behaviour itself does not have a magnitude in this model.

The Fogg behaviour Model operates on a straightforward binary premise, categorizing behaviour as either performed or not. The focus is not on the "magnitude" of the behaviour but rather on the frequency of engagement for habit formation. In this context, it is more effective to regularly engage in a simplified version of the target behaviour than to not perform it at all. Using walking as an example, some argue that it can be seen as a gradual behaviour with varying durations or frequencies. However, in Fogg's model, these are properties of a walk that can be adjusted. For instance, the length of the walk can be shortened to enhance the patient's ability to engage in it. This concept aligns well with BCIO, where the length of the walk may correspond to the BCI dose, which is a property of the BCI content. Our objective is to modify the BCI in a way that promotes the performance of the well-being behaviours.

While prior work focused on specific aspects of BCI for personalisation, our novelty is in leveraging both patient context and all BCI properties to predict whether a patient will perform the target behaviour. Moreover, we propose to utilize counterfactual examples with feature control to guide BCI personalisation.
A counterfactual example is a hypothetical input instance that is similar to the original instance, but with certain features modified in order to produce a different prediction from the machine learning (ML) model. For example model predicts that "John (age 67, male etc) given his current context (stressed, fatigued, outside of home, on Tuesday afternoon) will not perform a 45 minutes brisk walk if the VC sends him motivating reminder at this point." However, if we improve his ability to perform the BCI and change the walk duration to 30 minutes or to slower speed the model predicts that John will perform the suggested activity (See Figure \ref{personalisationSystem2}).

In this preliminary work we:
1) Identify BCI properties that can influence patients' Motivation, Ability, and Trigger (MAT) dimensions. We propose that these properties, along with patient context, can be used to predict whether the suggested behaviour will be performed.
2) Adapt the concept of counterfactual examples, commonly used in the field of explainable AI, to personalise BCI.
3) Utilize simulated data to design a study that enables the evaluation of the proposed personalisation system.



 \section{Related Work}

Kunzler \etal \cite{kunzler2019exploring} trained a Random Forest (RF) Classifier to predict participants' receptivity in-the-moment, specifically detecting when participants respond to a chatbot. By utilizing both participant context and their intrinsic characteristics features, they achieved an accuracy of 77\%. In our behaviour prediction model, we also incorporate these features and use RF. However, unlike Kunzler et al., our objective is not solely to predict when the patient responds to the notification and consequently adjust BCI delivery. We are also interested in determining whether they will perform the target behaviour as a result of the notification. Therefore, we additionally consider the Motivation and Ability dimensions of Fogg's model. These are affected by BCI content. 
For example, Mair et al. \cite{mair2022personalized} aimed to increase physical activity in older adults. In their study, participants were given the option to select their goal in terms of either time spent on performing physical activity or the number of steps walked. Each goal had three target ranges, allowing users to match the intervention dose to their ability. The participants received messages that aimed to motivate them (e.g., by highlighting the benefits of walking) or to enhance their ability to perform the activity (e.g., supporting activity planning). We note that these messages correspond to different Behaviour Change (BC) techniques. Following Mair et al., we consider BCI dose and BC technique as variables that can be personalised and we have incorporated them as input features into our behaviour prediction model.

 
\section{Methods}

 \begin{figure}
    \centering
    \includegraphics[width=12cm]{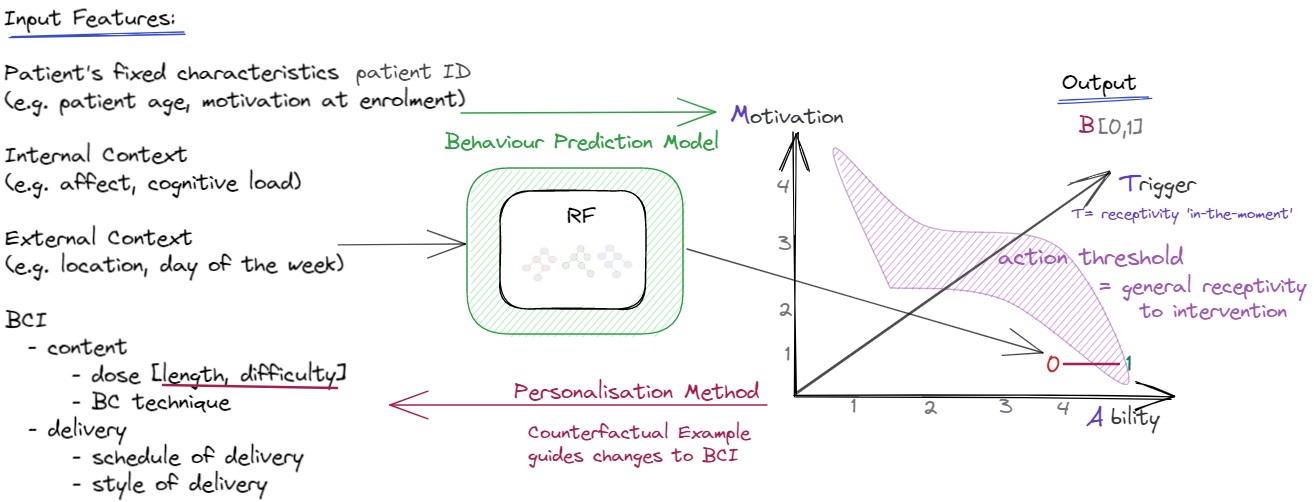}
    \caption{An example of utilizing counterfactual explanation with feature control to guide the personalisation of BCI. In the forward step, the model predicts that the behaviour will not be performed based on the current patient context and the selected BCI. We generate counterfactual examples and choose one that necessitates the fewest changes to move the patient above the action line. The counterfactual example demonstrates that we can retain the recommended activity type but need to adjust the BCI dose, such as suggesting a shorter walk.}
    \label{personalisationSystem2}
    \vspace{-5pt}
\end{figure}

\subsubsection{Behaviour Prediction Model}
\label{bpm}
 We trained an RF model to predict a patient's behaviour (i.e., the execution of a target activity framed as a binary classification problem) using their fixed characteristics (age, gender, motivation at enrollment), internal context (affect, cognitive load), external context (motion, location, time of day, day of the week), and BCI properties (activity type, dose, delivery schedule, message phrasing, and content) (See forward pass in Figure \ref{personalisationSystem2}). We use \textsf{scikit-learn} implementation of the RF with default parameters and balanced class weighting.

\subsubsection{Personalisation Method}
We utilize the DiCE (Diverse Counterfactual Explanation) framework \cite{mothilal2020dice} with feature control to identify specific changes needed in the BCI to increase the likelihood of patient adherence to it, i.e., performing the target behaviour (See the red backward arrow in Figure \ref{personalisationSystem2}) . We generate multiple counterfactual examples using a genetic algorithm, but for personalisation, we select the one that requires the fewest changes to the patient's context and BCI.To illustrate this, let's consider an example with John. If John were more motivated, relaxed, and well-rested, he would likely be willing to engage in a suggested 45-minute brisk walk. However, achieving these changes in his state might be more challenging compared to increasing his ability to perform the target behaviour by reducing the duration of the walk. Therefore, even though there are multiple counterfactual scenarios in which John takes a walk or engages in other well-being activities like yoga, we select the one that requires the minimal number of changes to his context and BCI, ideally altering only one BCI feature.

Feature control is a technique that ensures that generated counterfactual explanations satisfy certain constraints. For example, we might want to increase a patient ability to perform a target behaviour but we cannot change the fixed characteristics, e.g. age.  We allow changes to the features which relate to BCI but not to patient characteristics or their internal context.

\subsubsection{Evaluation Method} 
 There is a gap in health BCI research in terms of availability of public BCI datasets. To be able to evaluate an ML-based intervention personalisation approach prior to commencement of the clinical study it is necessary to create synthetic data. In our previous work \cite{lisowska2021personalized} we developed a simulator based on Fogg's model which computes MAT components based on multiple data items and uses equation \ref{eq:behaviour} to generate binary behaviour outputs (behaviour was performed or not). Here we simplify simulation so that each MAT dimension has an integer value from 0-4, depending only on model input features described in section \ref{bpm}.  Following equation \ref{eq:behaviour} the action thresholds should have values between 0 and 64. At the action threshold of 0, a behaviour is always performed, and at 64 -- never.
 
In this work, drawing inspiration from the findings of our previous survey study \cite{lisowska2021habit} and a pilot real-world well-being intervention involving healthy volunteers performing their selected target behaviour daily \cite{lisowska2022improve}, we simulate patients with different action thresholds to investigate the influence of these thresholds on the performance of the RF model.
Figure \ref{personalisationSystem2} depicts a non-linear action threshold. However, in the initial exploration, we generated simulated patients with linear thresholds. We acknowledge that this simplified setting may not fully capture the complexities of real-life scenarios, where initial increases in ability or motivation (from 0 to a small value) may have a larger impact on behaviour compared to increases from high to very high. Nevertheless, our objective was to assess whether the behaviour prediction model can be effectively trained using the selected input features in a small annotated data regime that reflects real-life interventions. If it fails to work in this simple setting, it is unlikely to perform well in more complex scenarios.

We also generate datasets with diverse numbers of simulated patients (between 1-100) with varying action thresholds (generated at random) to explore how the RF could benefit from learning across multiple patients. We train the model on all patients but use patient identifier as a feature to distinguish between patient or group of patients if it is necessary.

\section{Experiments}
The simulator is utilized to generate both training and testing examples. At the beginning of the intervention, the available data is very limited as it has not yet been gathered. To accurately reflect this situation we initiate the training process with a small training set and gradually increase it as the intervention progresses over time. For testing, we generate 400 samples per patient, while the number of samples for training varies from 2 to 32. It is important to note that during the intervention, we assume receiving one annotated sample per day from each individual, indicating whether the patient performed the target behaviour or not. We consider a maximum data collection period of one month. Consequently, the number of training data remains relatively low. However, this setup realistically reflects what we could expect in a real-life study.

\subsubsection{Learning to Predict Behaviour for Simulated Patients with Different Action Thresholds}
\label{DifferentAT}
\begin{figure}
    \centering
    \includegraphics[width=9cm]{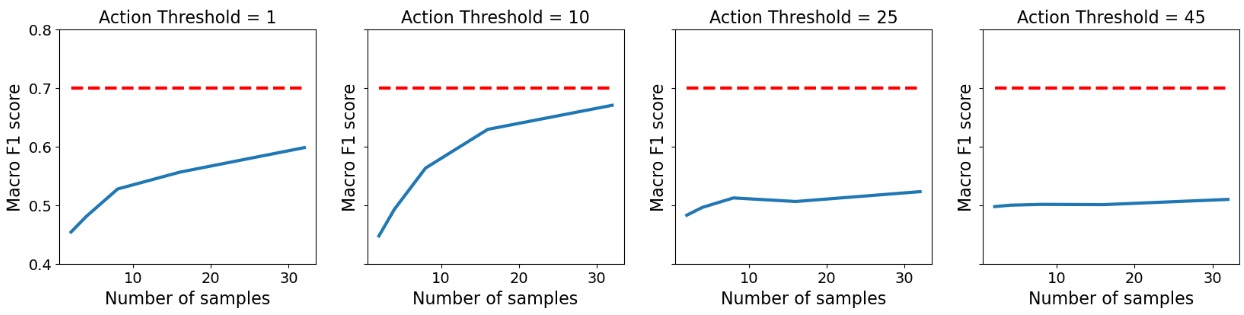} 
    \vspace{-5pt}
    \caption{Impact of action threshold on RF behaviour prediction performance.}
     \vspace{-5pt}
   \label{actionThreshold}
\end{figure}

We investigate how the action threshold impacts the number of samples required to train the RF model for behaviour prediction.  As shown in Figure \ref{actionThreshold},
for a receptive individual (action threshold = 10), RF needs around 30 samples to achieve approx 0.7 macro F1 score. When the activity threshold is low, we have only positive examples and RF is struggling to learn the context in which the behaviour is not performed. A similar problem arises when the activity threshold is set too high. 
 As expected in such cases, the model's effectiveness in training is compromised due to the limited availability of positive examples. However, it is noteworthy that the range of thresholds from which we can gather samples of performed behaviour is remarkably narrow. This suggests that the personalised model can only be effectively trained for a small subset of patients.

 From a machine learning perspective, this finding may not be unexpected, but it holds significant practical implications for real-life studies. Specifically, when comparing personalised approaches, it is vital to ensure that each study group exhibits a similar distribution of "general intervention receptivity." This issue becomes even more apparent in the subsequent experiment.



\subsubsection{Learning from Multiple Simulated Patients}
\label{multip}
\begin{figure}[t]
    \centering
    \subfloat [Predicting behaviour given context, BCI and participant ID features]{{\includegraphics[width=11cm]{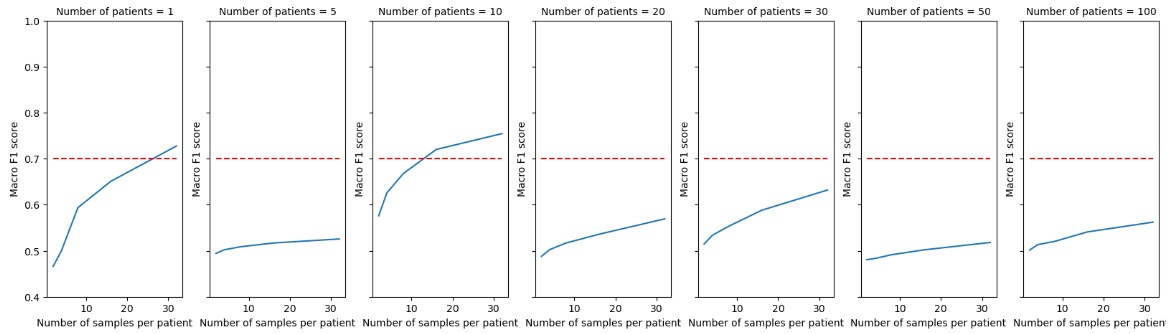}} \label{raw}}
    
        \subfloat[Predicting behaviour given MAT values and participant ID ]{{\includegraphics[width=11cm]{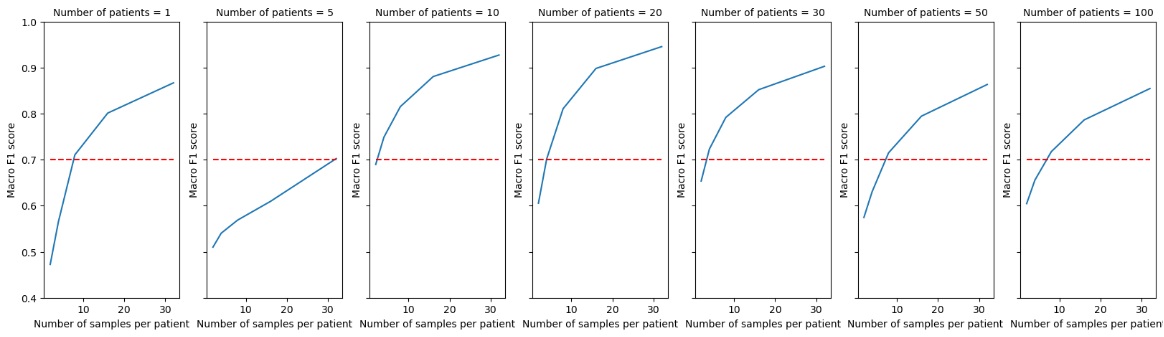}} \label{matin}}
        
          \subfloat[Predicting behaviour, motivation, trigger, ability given context and BCI]{{\includegraphics[width=11cm]{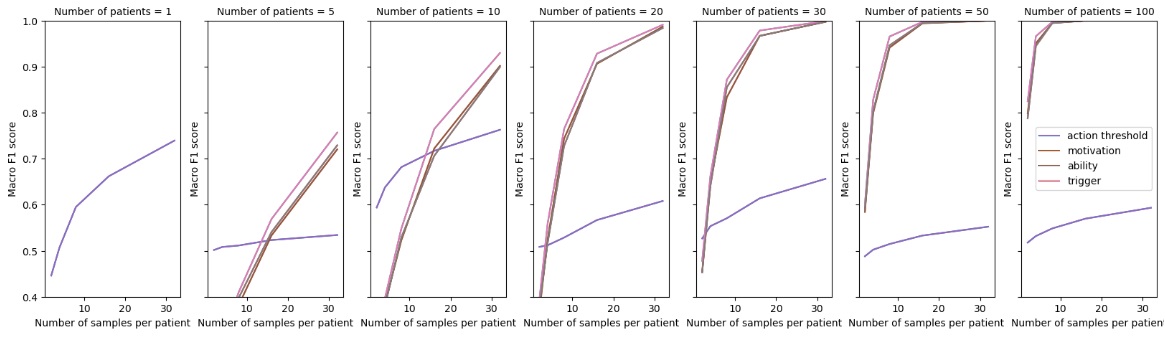}}\label{matout}} 
    \vspace{-5pt}
    \caption{Performance of RF model for varying number of simulated participants with different action thresholds. The \% of participants with action threshold below 40 for plots from left (1) to right (100) is: 100, 40, 80, 65, 70, 52.}
    \vspace{-19pt}
\end{figure}

Figure \ref{raw} shows that increasing number of samples through increasing number of patients in a study might not necessarily lead to better learning of behaviour prediction because patients would likely have different general receptivity levels. The proportion of patients with higher general receptivity to intervention might be more predictive of model performance than the total number of simulated patients. For example, learning from a smaller group of 10 patients (with 30 samples per patients), where 80\% have an action threshold below 40, leads to a higher macro F1 score for the RF model compared to learning from a larger group of 100 patients, where only 52\% have an action threshold below 40.

\subsubsection{Adding Supervision for Enhanced Learning}
\label{sup}
 We use the MAT values from the simulation as class output labels for a multi-class RF classifier. 
 There might be variables that commonly contribute to one of MAT dimensions, as assumed in our simulation. For instance, activity difficulty and the patient's physiological state may both contribute to patients ability to perform the target activity. Therefore, instead of directly predicting behaviour, it may be easier for the model to learn where the patient stands on each axis separately. This is especially true when dealing with multiple patients, as shown in Figure \ref{matout}. Hence prediction of MAT could be an intermediate step in intervention personalisation. The MAT values might be subsequently fed alongside the patient ID to the next RF model to learn to predict the behaviour for each individual (See Figure \ref{matin}). 

\section{Discussion}

Our experiments revealed that RF struggles to directly predict whether a patient performs the target behaviour based on input features alone. Therefore, we revise the Behaviour Prediction Model to be a two-step system: 1) Predicting the MAT and 2) Predicting if the patient is above or below the action threshold. In this revised system to personalise the intervention, we use DiCE twice: 1) Identify the most effective MAT dimension to target for a given patient in a given context, and 2) Identify BCI properties that should be modulated to move the patient forward on the selected MAT axis. This approach assumes that while patients may differ in their general receptivity to interventions, there may be common factors that influence their motivation, ability, or receptivity in the moment.

The main limitation of our experiments is that the MAT values are generated by the simulated system. 
In a real-world intervention, MAT annotation would be obtained from user interactions with the VC. Trigger values could be assigned based on user responsiveness to notifications or ignored messages. Additional information about patients' ability and motivation might come from enrollment questionnaires, direct questions after the performance of the target behaviour, or ratings of activities and motivational messages. For activities with demonstration videos (e.g., 'Tai Chi' lesson), we can approximate the patient's ability by measuring how long the video was played. The capture of this information would need to be planned for already at the BCI design stage.

In our future work, we aim to explore methods for capturing additional supervision information more seamlessly, without burdening patients, and integrating data collected at various frequencies to drive BCI personalisation.

\section{Conclusion}
In this preliminary work, our objective is to design a study that effectively tests our personalised system. Specifically, we aim to determine the optimal number of patients required in the study to train a behaviour prediction model effectively. Additionally, we investigate which information about the patient and intervention needs to be leveraged throughout the study. Through simulating diverse patient profiles, we analyze how individual receptivity to the intervention impacts the training of the machine learning model. Our findings highlight that patients' receptivity to intervention has a stronger influence on model performance than the number of patients in the study. This emphasizes the importance of ensuring a similar distribution of "general intervention receptivity" in each study group when comparing personalised approaches. Moreover, our findings demonstrate that incorporating intermediate representation, which considers individual MAT components, contributes to the successful learning of behaviour prediction and subsequent personalisation of BCI. In practice, this necessitates inferring measures of patients' ability and motivation from their interactions with the mHealth application or collecting such data during the study. Lastly, to facilitate patients in performing the target behaviour, we propose employing the DiCE method to generate potential strategies for modulating the BCI.

\subsubsection*{Acknowledgments}
This work has received funding from the EU's Horizon 2020 research and innovation programme under grant agreement No 875052.

\bibliographystyle{unsrt}

\vspace{-6pt}
\bibliography{ref}

\end{document}